\documentclass[conference]{IEEEtran}
\IEEEoverridecommandlockouts
\usepackage{amsmath,amssymb,amsfonts}
\usepackage{algorithm}
\usepackage{algorithmic}

\usepackage{graphicx}
\usepackage{booktabs}
\usepackage{url}
\usepackage{xcolor}
\usepackage[hidelinks,breaklinks=true]{hyperref}
\newif\ifrevision\revisionfalse
\newcommand{\rev}[1]{\ifrevision{\color{blue}#1}\else#1\fi}
\newenvironment{revblock}{\ifrevision\color{blue}\fi}{}

\begin{document}

\title{Where Is the Truth When AI Agents Disagree?\\{\large Dependence-Aware Belief Propagation for Unsupervised Truth Discovery\\[2pt] [Extended Version]}}

\author{\IEEEauthorblockN{Laure Berti-Equille}
\IEEEauthorblockA{IRD, ESPACE-DEV\\
Montpellier, France\\
laure.berti@ird.fr}}

\IEEEpubid{\makebox[\columnwidth]{\parbox{\columnwidth}{\scriptsize\copyright~2026 IEEE. Personal use of this material is permitted. Permission from IEEE must be obtained for all other uses, in any current or future media, including reprinting/republishing this material for advertising or promotional purposes, creating new collective works, for resale or redistribution to servers or lists, or reuse of any copyrighted component of this work in other works. Accepted at the 2026 IEEE International Conference on Big Data (BigData 2026).}}\hspace{\columnsep}\makebox[\columnwidth]{}}

\maketitle
\IEEEpubidadjcol

\begin{abstract}
\begin{revblock}
LLM agents often give conflicting answers to the same question, and many deployed
agents are variants of one base model, so their errors are correlated. We study
truth discovery among such agents: inferring the true answer and each agent's
reliability without labels. Classical truth-discovery and crowdsourcing estimators
reward an agreeing bloc and can fall below majority vote. We propose BP-TD, which
adds per-agent decorrelation weights to mean-field variational inference on a
one-coin Dawid--Skene model, so that agents sharing a base model count together as
about one vote. Its first iteration is exactly a weighted vote; later iterations
estimate agent reliability. With $13$ local agents on MMLU, BP-TD beats majority vote
without any labels ($0.716$ vs.\ $0.624$ on $160$ questions; $0.707$ vs.\ $0.674$ on
$10{,}000$) and ranks agents by their true accuracy, which classical estimators
invert. The gain depends on the pool: over $2{,}852$ sub-pools of four benchmarks,
decorrelation helps when the largest correlated bloc is weaker than the other agents
and hurts when it is the strongest, and we found no label-free statistic that tells
the two cases apart. Labels change the picture quickly: with $20$ labelled questions,
selecting the single best agent is more accurate than every aggregator we tested on
MMLU, ARC and TruthfulQA.
\end{revblock}
\end{abstract}


\section{Introduction}
Ask the same factual question to several large language model (LLM)
agents and you will often get several different answers: on a
multiple-choice item, one model answers~A, another~B, a third abstains.
As LLM agents are wired together into pipelines, debates, and
mixtures, this disagreement is no longer a curiosity but an operational
problem---a system must commit to \emph{one} answer, ideally the true
one, and ideally without paying for ground-truth labels at every step.
Deciding which answer to trust is exactly \emph{truth discovery} (TD): a
data-quality problem with two decades of theory for conflicting Web
sources and crowd workers
\cite{yin2008truthfinder,dong2009integrating,li2014resolving,dawid1979maximum}.
Whether that theory transfers to AI agents, and \emph{when} it does, has
not been characterized.

\textbf{The gap.} Classical TD rests on a structural assumption:
sources are \emph{conditionally independent} given the truth, so that
agreement is evidence of correctness and reliability can be estimated
from how often a source sides with the emerging consensus. This
assumption is routinely violated by AI agents for a reason specific to
them: deployed agents are frequently \emph{variants of the same base
model}---prompt personas, self-consistency samples, or retrieval
wrappers around one LLM---so their errors are strongly \emph{correlated}.
Reliability-by-agreement then rewards a correlated bloc that may be
confidently and unanimously \emph{wrong}, and the standard TD machinery
inverts: it can rank the agreeing-but-wrong majority as most reliable.

\begin{revblock}
\textbf{Our approach.} We propose \textbf{BP-TD}, which places a per-agent
decorrelation weight $w_a$ inside mean-field variational inference on a one-coin
Dawid--Skene model. Agents that share a base model, or that agree with each other far
more than chance, share one vote. The weights come either from provenance or from
answer co-occurrence, with no labels. The first iteration of BP-TD is exactly a
$w_a$-weighted plurality vote; later iterations add per-agent reliability estimates.
We test the method on $13$ local agents and four benchmarks, against classical,
dependence-aware and concurrent LLM-aggregation methods, and we vary the agent pool to
find out when decorrelation helps and when it hurts.

\textbf{Contributions.}
\begin{itemize}
\item[\textbf{C1.}] \emph{Why classical TD fails on LLM agents.} Reliability
estimators that reward agreement rank a correlated bloc of weak agents as the most
reliable and fall below majority vote. The same implementations work on independent
agents, so the failure comes from the dependence, not from the estimators.
\item[\textbf{C2.}] \emph{A decorrelation method and what each part does.} BP-TD
discounts dependence groups found from provenance or from answer co-occurrence. We
show that its first iteration is a weighted vote, separate the effect of the weights
from that of iteration and of the threshold choice, and give label-free settings of
its one hyperparameter.
\item[\textbf{C3.}] \emph{When agent truth discovery helps.} Decorrelation helps when
the largest correlated bloc is weaker than the other agents and hurts when it is the
strongest; this holds across thousands of sub-pools of four benchmarks and in a
controlled stress test. The gain on MMLU survives at $10{,}000$ questions. Twenty
labels are worth more than any aggregation we tested, so BP-TD is useful when labels
cannot be had and the pool's largest bloc is not its strongest.
\end{itemize}
The code, the cached agent answers with provenance, and every script behind the
numbers below are released.
\end{revblock}

\section{Related Work}\label{sec:related}
\textbf{Classical truth discovery.} A large body of work jointly estimates
source trust and claim belief by iterating between the two. Iterative
fixed-point methods include TruthFinder~\cite{yin2008truthfinder} and the
Investment and Sums/Hubs--Authorities schemes~\cite{pasternack2010knowing},
while CRH and CATD~\cite{li2014resolving} cast the problem as joint
optimization with confidence-aware weighting; earlier fixed-point
corroboration~\cite{galland2010corroborating}, correlation-aware
fusion~\cite{pochampally2014fusing}, and knowledge fusion over machine-extracted
triples~\cite{dong2014knowledgefusion} broaden the family (surveyed
in~\cite{li2016survey,bertiequille2015veracity,bertiequille2016crossmodal}), and controlled evaluations find no estimator dominates
across source regimes~\cite{waguih2014truthdiscovery,bertiequille2015ensembling}. These methods are unsupervised
and effective when sources err independently, but they provide no account of
\emph{why} a method should be expected to help on a given workload---the
regime question we make central.

\textbf{Crowdsourcing aggregation.} Latent-variable models treat each source as
a noisy annotator and infer a confusion structure: Dawid--Skene estimates
per-annotator error rates by EM~\cite{dawid1979maximum}, and LTM extends this
with a generative Bayesian treatment of truth and reliability~\cite{zhao2012bayesian};
further variants estimate annotator ability and item difficulty
(GLAD~\cite{whitehill2009glad}), infer confusability by minimax
entropy~\cite{zhou2012minimax}, or aggregate by message passing---order-optimal in~\cite{karger2011iterative},
variational/BP in~\cite{liu2012variational}--- with a uniform benchmark
of seventeen such methods~\cite{zheng2017truthinference}.
Belief propagation over the Dawid--Skene model is itself established and provably
optimal under \emph{independent} annotators~\cite{ok2016optimal}; our
contribution is the dependence-aware decorrelation layer that keeps inference
valid when annotators---here, agents---are \emph{correlated}, which that line
does not address.
Our message-passing update reduces to a one-coin Dawid--Skene model when
agent weights are uniform; the contribution is not the latent-variable
machinery but the dependence-aware weighting layered on top of it.

\textbf{Source dependence and copying.} A key assumption of the above is
conditional independence of sources, which copying violates. Dong et
al.~\cite{dong2009integrating} detect copying among Web sources from shared
\emph{false} values and down-weight copied evidence, with global detection of
complex copying networks across many sources~\cite{dong2010global}; in
crowdsourcing, spectral
and coupled-confusion methods handle correlated annotators beyond independent
Dawid--Skene~\cite{zhang2014spectral}. Large-scale evidence confirms the
phenomenon directly for LLMs: higher-capability and same-provider models make
\emph{correlated} errors even across architectures~\cite{kim2025correlated}.
Our setting is the agent
analogue: agents that share a base model exhibit correlated errors by
construction. We adapt this intuition with family- and agreement-based
discounting so that a correlated bloc counts as roughly one \emph{effective} vote,
and we show that a shared-error signal---the direct copy-detection
analogue---degrades when the dependent bloc \emph{is} the majority.

\textbf{LLM ensembling, self-consistency, and debate.} Recent work aggregates
multiple LLM samples or agents---self-consistency~\cite{wang2023selfconsistency},
multi-agent debate~\cite{du2024debate}, mixture-of-agents style
ensembling~\cite{wang2024mixtureofagents}, learned pairwise-ranking
fusion~\cite{jiang2023llmblender}, sampling-and-voting at scale~\cite{li2024moreagents},
juries of diverse judges~\cite{verga2024juries}, and LLM-as-a-judge evaluation~\cite{zheng2023judging}, often leaning on self-reported model
confidence~\cite{kadavath2022knowwhat}---generally assuming a cooperative crowd
whose members add complementary signal. Reference-free aggregation of LLM judges
has been studied directly~\cite{sun2025skill}, and recent analyses question the
diversity assumption: mixing distinct models can underperform ensembling a single
strong one~\cite{li2025rethinkingmoa}, and correlated judges cap panel gains
regardless of the aggregation rule~\cite{kohli2026nine}. \rev{Concurrent work
aggregates \emph{dependent} LLM judges with second-order answer statistics (ISP and
OW~\cite{ai2025beyondmajority}), with Ising models of judge
dependence~\cite{balasubramanian2026ising}, or by separating quality from shared
confounders (CARE~\cite{zhao2026care}). The Ising aggregator is defined for binary
labels only, so we use a mean-field surrogate.}

\rev{\textbf{What is new here.} BP-TD builds on one-coin Dawid--Skene
inference~\cite{dawid1979maximum,liu2012variational}, discounting of dependent
sources~\cite{dong2009integrating,pochampally2014fusing}, and grouping of correlated
voters (our ClusterVote baseline). What differs with BP-TD is the dependence signal it
uses: provenance or raw answer co-occurrence,
never the current truth estimate, which we show is circular when the dependent bloc
is the majority. It differs from cluster voting in keeping soft per-agent weights
inside a reliability model rather than collapsing each group to one hard vote; on
MMLU, with identical groups, this is worth $+0.043$ (Sec.~\ref{sec:res-why}). The
main empirical finding---decorrelation helps when the largest correlated bloc is
weaker than the other agents and hurts when it is the strongest---is not specific to
BP-TD: any rule that discounts a group of agreeing agents gives up that group's votes,
which costs accuracy exactly when the group is right.}

\textbf{Signed graphs.} Balance theory and signed bipartite graph neural
networks~\cite{huang2021signed} exploit the \emph{sign} of relations for
representation learning. We use signed agreement for the opposite purpose: not
to embed agents, but to detect correlated blocs and \emph{decorrelate} their
votes during inference, keeping the aggregator unsupervised and label-blind.

\section{Problem Formulation}\label{sec:problem}
We are given a set of \emph{agents} $A=\{a_1,\dots,a_m\}$ (here, LLM-based
answerers) responding to a set of \emph{questions} $Q=\{q_1,\dots,q_n\}$, where
each question $q$ has a known, finite \emph{answer set} $V_q$ of size
$|V_q|$. For each $(a,q)$ pair, agent $a$ either asserts a single value
$y_{aq}\in V_q$ or abstains; let $A_q\subseteq A$ denote the agents that answer
$q$. We seek, with \emph{no gold labels at inference time}, (i) the latent truth
$t_q^\star\in V_q$ for every question, and (ii) a scalar \emph{reliability}
$r_a\in[0,1]$ for every agent, used as a relative ranking score, not a
calibrated probability of correctness (BP-TD recovers the ranking, not the magnitude).

\begin{figure}[t]
  \centering
  \includegraphics[width=\linewidth]{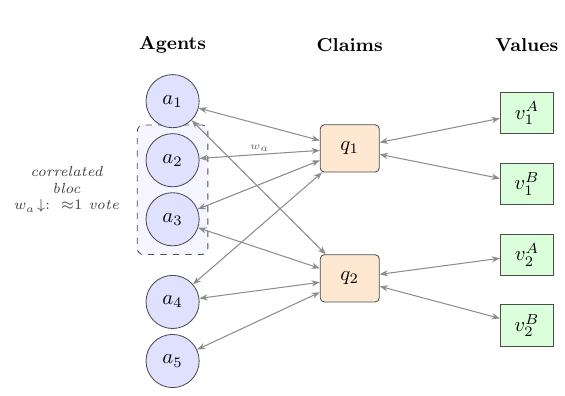}
  \caption{\rev{The agent--question--value graph and the two messages of BP-TD. An
    edge links an agent to the value it asserts on a question. Truth update
    (agents$\to$questions, Eq.~1): each agent sends its weighted vote
    $w_a\,\mathbb{E}\log r_a$ to the value it chose. Reliability update
    (questions$\to$agents, Eqs.~2--3): each agent receives the current belief in the
    values it asserted, which updates its Beta reliability. Dashed box: a dependence
    group (agents on one base model); each member's vote is scaled by $w_a$, so the
    group counts as about one vote.}}
  \label{fig:graph-schematic}
\end{figure}

We represent the observations as a tripartite \emph{agent--question--value}
graph (Fig.~\ref{fig:graph-schematic}): nodes are agents, questions, and candidate values, and an edge $(a,q,v)$
exists iff $y_{aq}=v$. This makes the inference structure explicit---belief flows
from agents to questions (which value is likely true) and back from questions to
agents (how reliable each agent is).

\textbf{The dependence problem.} Classical truth discovery and the Dawid--Skene
model~\cite{dawid1979maximum} assume agents are \emph{conditionally independent}
given the truth, so that agreement is evidence of correctness. This assumption is
violated for LLM ensembles: agents that share a base model, a prompt template, or
a retrieval corpus produce \emph{correlated errors}, and a single such bloc can
manufacture a spurious majority. Our goal is therefore aggregation that
\emph{down-weights} dependent agents so a correlated bloc contributes roughly one
\emph{effective} vote.

\section{Method: BP-TD}\label{sec:method}
BP-TD combines (i) variational message passing on a one-coin Dawid--Skene factor
graph with (ii) a per-agent \emph{decorrelation weight} $w_a$ that encodes how
much an agent's vote should count given its dependence on others.
\rev{On terminology: the inference is mean-field variational message passing
(coordinate ascent on the evidence lower bound), not sum-product belief propagation.
We keep the name BP-TD because the updates are messages passed along the edges of the
agent--question--value graph of Fig.~\ref{fig:graph-schematic}, and loopy BP and
mean-field updates share this structure; the estimator itself is the mean-field one
below.}

\subsection{Variational message passing}
We adopt a \emph{one-coin} observation model: with probability $r_a$ agent $a$
emits the true value, and otherwise emits a value drawn uniformly from the
$|V_q|-1$ wrong alternatives. Placing a $\mathrm{Beta}(\alpha_0,\beta_0)$ prior on
each $r_a$, mean-field variational inference factorizes the posterior over the
question beliefs $b_q(\cdot)$ and the reliabilities $r_a$, and alternates two
closed-form updates:
\begin{align}
\log b_q(v) &\propto \textstyle\sum_{a\in A_q}
   w_a\big[\mathbf{1}[y_{aq}{=}v]\,\mathbb{E}\log r_a \nonumber\\
   &\quad + \mathbf{1}[y_{aq}{\neq}v]\big(\mathbb{E}\log(1{-}r_a)-\log(|V_q|{-}1)\big)\big],\\
\alpha_a &= \alpha_0 + \textstyle\sum_q b_q(y_{aq}), \\
\beta_a  &= \beta_0 + \textstyle\sum_q \big(1-b_q(y_{aq})\big).
\end{align}

\textbf{Intuition.} The \emph{truth update} (Eq.~1) is a weighted vote: each
answering agent adds $w_a\,\mathbb{E}\log r_a$ to its asserted value and a smaller
$w_a\,\mathbb{E}\log(1{-}r_a)$ to the $|V_q|-1$ competitors, so reliable agents push
their answer harder while $w_a$ shrinks the influence of dependent agents. The
\emph{reliability update} (Eqs.~2--3) is conjugate Beta accounting: $\alpha_a$
accumulates the belief mass on $a$'s chosen value ($\sum_q b_q(y_{aq})$) and
$\beta_a$ the complementary mass, so $\mathbb{E}[r_a]=\alpha_a/(\alpha_a+\beta_a)$
rises for agents that agree with confident consensus. The two updates form a
soft, EM-style loop; setting $w_a\!\equiv\!1$ removes decorrelation and recovers
one-coin Dawid--Skene / SSTF-VI.

\rev{\textbf{One pass is a weighted vote.} At initialization every agent has the
same prior reliability, so $\mathbb{E}\log r_a$ and $\mathbb{E}\log(1{-}r_a)$ are
the same constants for all $a$. Eq.~(1) then gives
$\log b_q(v)=c_q+\kappa\sum_{a\in A_q} w_a\mathbf{1}[y_{aq}{=}v]$ with
$\kappa=\mathbb{E}\log r-\mathbb{E}\log(1{-}r)+\log(|V_q|{-}1)>0$, so the first
truth update ranks values exactly as the $w_a$-weighted plurality vote. A single
pass ($T{=}1$) therefore predicts the weighted vote, and with $w_a\equiv1$ it
predicts majority vote. Reliability estimation affects the predicted answers only
from the second iteration on. We report both the one-pass weighted vote and the
iterated estimator, and state the iteration count of every BP-TD row.}

\rev{\textbf{Why $w_a$ tempers the likelihood.} The weights are not part of the
generative model; they enter Eq.~(1) as exponents on each agent's likelihood term.
This is a composite (power) likelihood: when $k$ agents are near-copies, the product
of their $k$ likelihood terms counts the same evidence $k$ times, and raising each
to the power $1/k$ restores roughly one copy's worth of evidence. The update for
$b_q$ stays a closed-form coordinate-ascent step for this tempered objective, and the
reliability update (Eqs.~2--3) is unchanged. A generative model of dependence
(e.g.\ a latent bloc variable) would replace this heuristic correction; we leave
that to future work.}

\subsection{Dependence-aware decorrelation}
The decorrelation weight $w_a\in(0,1]$ is the core contribution: it discounts an
agent in proportion to the size of the dependent bloc it belongs to, so that a
$k$-agent bloc contributes $\sum_{a\in\text{bloc}}w_a\approx 1$ effective vote
rather than $k$. We compare three ways to define the blocs, ordered by how much
external information they require:
\begin{itemize}
\item \textbf{family} (structural). Agents sharing a base-model family form one
cluster; $w_a=1/|\text{family}(a)|$. Requires only provenance (which base model
produced each agent), no observed answers. \emph{Cheapest, label-free, fixed.}
\item \textbf{cooccur} (empirical, truth-free). Discount each agent by its number
of above-threshold neighbors in the signed co-occurrence graph---how often two
agents give the \emph{same} answer, regardless of whether it is correct:
$w_a=1/\big(1+|\{b:\mathrm{agree}(a,b)\ge\theta\}|\big)$, so an agent that agrees
with $k{-}1$ others counts as $\approx$$1/k$ of a vote. Uses the answer matrix
but \emph{not} any estimate of truth. \emph{Adapts to behavior without trusting
the current belief.}
\item \textbf{shared\_error} (truth-aware). Following copy detection on shared
\emph{false} values~\cite{dong2009integrating}, cluster agents that agree on the
same \emph{likely-wrong} answers, where ``wrong'' is read off the current beliefs
$b_q$; recomputed each iteration. \emph{Most principled in
principle---correlated mistakes are the strongest dependence signal.}
\end{itemize}

\textbf{Why \texttt{shared\_error} fails when \texttt{family}/\texttt{cooccur}
succeed.} \texttt{shared\_error} defines dependence relative to the inferred
truth: two agents are dependent if they share errors. But when the dependent bloc
\emph{is} the majority, it \emph{defines} the belief $b_q$---the bloc's answers
look correct, so its members appear to share no errors and escape discounting.
The measure is thus circular precisely in the regime it is meant to fix: it
trusts the consensus to decide who is correlated, but the correlation is what
corrupts the consensus. By contrast, \texttt{family} (provenance) and
\texttt{cooccur} (raw agreement) never consult the bloc-defined truth, so they
detect and discount a dominant bloc regardless of whether it is currently winning
the vote---which is why they are robust (Sec.~\ref{sec:results}). The
signed-agreement view underlying \texttt{cooccur} connects to balance theory on
signed graphs~\cite{huang2021signed}, used here to decorrelate rather than to
embed.

\subsection{Algorithm and complexity}
Algorithm~\ref{alg:bptd} summarizes BP-TD. Let $|E|=\sum_q|A_q|\le n m$ be the
number of asserted answers. Each iteration costs $O(|E|\cdot \bar{|V|})$ for the
truth update (Eq.~1) plus $O(|E|)$ for the reliability update (Eqs.~2--3), and the
dependence step costs $O(m^2)$ for \texttt{cooccur}/\texttt{shared\_error}
(pairwise agent agreement) or $O(m)$ for \texttt{family}. Per iteration this is
$O(n\,m\,\bar{|V|}+m^2)$, linear in the data and quadratic only in the (small)
agent count; total cost is that times the number of iterations $T$.

\begin{algorithm}
\caption{BP-TD: dependence-aware variational truth discovery}
\label{alg:bptd}
\small\raggedright
\begin{algorithmic}[1]
\STATE \textbf{Input:} answers $\{y_{aq}\}$, value sets $\{V_q\}$, prior
       $(\alpha_0,\beta_0)$, dependence mode $\in\{\texttt{family},
       \texttt{cooccur},\texttt{shared\_error}\}$, iterations $T$,
       optional seed set $S$
\STATE \textbf{Init:} $\mathbb{E}[r_a]\!\leftarrow\!\alpha_0/(\alpha_0+\beta_0)$;
       $b_q\!\leftarrow$ uniform on $V_q$ \COMMENT{clamp $b_q$ for $q\in S$ to seeds}
\STATE Compute $w_a$ \COMMENT{\texttt{family}/\texttt{cooccur} precomputed once;
       \texttt{shared\_error} recomputed in the loop}
\FOR{$\tau=1$ \TO $T$}
  \FORALL{questions $q\notin S$}
     \STATE update $b_q(v)$ via Eq.~(1) and normalize
  \ENDFOR
  \FORALL{agents $a$}
     \STATE update $\alpha_a,\beta_a$ via Eq.~(2);
     $\mathbb{E}[r_a]\!\leftarrow\!\alpha_a/(\alpha_a+\beta_a)$
  \ENDFOR
  \STATE \rev{smooth $\mathbb{E}[r_a]$ over the signed agreement graph (weight $\lambda$)}
  \IF{mode $=$ \texttt{shared\_error}}
     \STATE recompute $w_a$ from current $b_q$
  \ENDIF
\ENDFOR
\STATE \textbf{Output:} $t_q^\star=\arg\max_v b_q(v)$ and reliabilities
       $\mathbb{E}[r_a]$
\end{algorithmic}
\end{algorithm}

\subsection{Variants}
\textbf{One-pass} runs a single truth+reliability update ($T{=}1$), the cheapest
setting, \rev{and is exactly the $w_a$-weighted vote shown above}. \textbf{Full loopy}
iterates Eqs.~(1)--(2) to convergence\rev{, which is the default and the variant behind
every ``BP-TD'' row unless a row is marked one-pass. Two further steps run in each
iteration. The truth messages are damped ($b_q\leftarrow0.7\,b_q^{\text{new}}+0.3\,
b_q^{\text{old}}$). The reliability means are then smoothed over the signed agreement
graph $W$ (agreement minus disagreement rate per agent pair):
$\bar r_a\leftarrow\tfrac12+(1{-}\lambda)(\bar r_a{-}\tfrac12)+\lambda\sum_b
W_{ab}(\bar r_b{-}\tfrac12)/\sum_b|W_{ab}|$ with $\lambda=0.4$, so an agent that
agrees with reliable agents and disagrees with unreliable ones gains reliability.
Iteration stops when the largest belief change is below $10^{-4}$ (at most $100$
iterations); the mean iteration count per dataset and variant is given with each result table ($7$--$21$ for most rows; $93$ on SciQ when $\theta{=}0.2$ is selected, where beliefs converge slowly near unanimity). Removing the
smoothing ($\lambda=0$) changes full-set MMLU accuracy by less than $0.013$
(Sec.~\ref{sec:res-why}).} \textbf{Semi-supervised seeding} clamps a fraction
$S$ of questions to known answers, which propagates through the reliability update
to calibrate the rest; we evaluate only on $Q\setminus S$.

\section{Experimental Setup}\label{sec:setup}
We evaluate BP-TD under a deliberately conservative, leakage-free protocol and
report every operating point---including those unfavorable to our method.

\paragraph{Agent pool}
\rev{The pool has $13$ answering agents built on $5$ base models, all served locally
via Ollama (no external API calls). Table~\ref{tab:agents} lists every agent with its
base model and accuracy. The pool combines (i)~six direct-prompting agents
(\texttt{qwen2.5:14b}, \texttt{qwen2.5:7b}, \texttt{llama3.2}, \texttt{mistral},
\texttt{phi3.5}, \texttt{gemma2}; the two \texttt{qwen2.5} sizes share one base model); (ii)~two self-consistency agents
(five sampled chains, majority over chains); (iii)~three persona agents; and (iv)~two
retrieval-augmented (RAG) agents. Terminology: an \emph{agent} is one answering
configuration; its \emph{base model} (or \emph{family}) is the underlying LLM; a
\emph{dependence group} is a set of agents that BP-TD discounts together, defined
either by base model (\texttt{family}) or by answer agreement (\texttt{cooccur}).
Three further configurations were run and are excluded because they never produced a
parseable answer: two \texttt{deepseek-r1} reasoning agents returned only truncated
\texttt{<think>} traces at the harvest token budget, and an external CLI agent
(OpenClaw) had no headless command configured. OpenClaw is kept in the released
answer files as an all-abstain row; it casts no vote and does not change any result.}
The pool is \emph{intentionally correlated}: the mean \emph{signed} pairwise agreement
(agree minus disagree per co-answered question, $\in[-1,1]$) is $0.33$--$0.35$ on
MMLU and TruthfulQA (raw agree-rate $\approx$$0.66$). \rev{It is also unbalanced in a
way that matters for the results: on MMLU the largest dependence group ($6$
\texttt{llama3.2} agents) is the least accurate ($0.43$--$0.49$), while the second
group ($4$ \texttt{qwen2.5} agents) is the most accurate ($0.76$--$0.83$). Some
agents are exact or near copies: the three \texttt{qwen2.5:7b} agents have the same
accuracy on MMLU, TruthfulQA and ARC. Sec.~\ref{sec:pools} varies this composition.}

\begin{table}[t]
\centering
\caption{\rev{Agent pool: base model, agent type, and accuracy per dataset (measured on
gold, on answered questions). OpenClaw never answers.}}
\label{tab:agents}
\resizebox{\columnwidth}{!}{%
\begin{tabular}{lllcccc}
\toprule
Agent & Base model & Type & MMLU & SciQ & TQA & ARC \\
\midrule
\texttt{direct:qwen2.5:14b} & qwen2.5 & direct & 0.825 & 0.988 & 0.810 & 0.919 \\
\texttt{direct:qwen2.5:7b} & qwen2.5 & direct & 0.762 & 0.981 & 0.740 & 0.906 \\
\texttt{rag:qwen2.5:7b} & qwen2.5 & RAG & 0.762 & 0.975 & 0.740 & 0.906 \\
\texttt{sc:qwen2.5:7b} & qwen2.5 & self-cons. & 0.756 & 0.981 & 0.740 & 0.906 \\
\texttt{sc:llama3.2:latest} & llama3.2 & self-cons. & 0.487 & 0.950 & 0.520 & 0.744 \\
\texttt{rag:llama3.2:latest} & llama3.2 & RAG & 0.472 & 0.969 & 0.520 & 0.731 \\
\texttt{direct:llama3.2:latest} & llama3.2 & direct & 0.465 & 0.950 & 0.531 & 0.725 \\
\texttt{persona:cautious} & llama3.2 & persona & 0.459 & 0.950 & 0.580 & 0.713 \\
\texttt{persona:expert} & llama3.2 & persona & 0.444 & 0.956 & 0.540 & 0.694 \\
\texttt{persona:skeptic} & llama3.2 & persona & 0.434 & 0.925 & 0.580 & 0.694 \\
\texttt{direct:mistral:7b} & mistral & direct & 0.631 & 0.944 & 0.560 & 0.762 \\
\texttt{direct:phi3.5:latest} & phi3.5 & direct & 0.625 & 0.981 & 0.500 & 0.850 \\
\texttt{direct:gemma2:2b} & gemma2 & direct & 0.613 & 0.931 & 0.440 & 0.694 \\
\texttt{openclaw} & openclaw & CLI & --- & --- & --- & --- \\
\bottomrule
\end{tabular}}
\end{table}

\paragraph{Datasets}
\begin{revblock}
Four closed-book multiple-choice benchmarks, all answered by the same pool.
\textbf{MMLU} ($160$ questions, stride-sampled across subjects) has high
disagreement among agents. \textbf{SciQ} ($160$) is grade-school science on which
almost every agent is right. \textbf{TruthfulQA-MC1} ($100$) is adversarial by
construction: popular misconceptions are offered as wrong answers. \textbf{ARC-Challenge}
($160$) is grade-school reasoning with moderate disagreement. For scale we use two
further MMLU harvests of $5{,}000$ and $10{,}000$ questions answered by the same $13$
agents served with vLLM in half precision (Sec.~\ref{sec:res-scale}). The SciQ and
TruthfulQA harvests cover their full question sets only at this size; we did not
harvest them at larger scale. A BoolQ pilot is reported in
Appendix~\ref{app:boolq}; it is not a valid closed-book test because $11$ of the $13$
agents did not see the passage the task requires.
\end{revblock}

\paragraph{Baselines}
\begin{revblock}
Majority vote (MV); the classical TD and crowdsourcing estimators Dawid--Skene,
TruthFinder, Investment, Sums, CRH, CATD, LTM, WeightedMV and SSTF-IV/VI
\cite{li2014resolving,yin2008truthfinder,dawid1979maximum,pasternack2010knowing,zhao2012bayesian};
the dependence-aware ACCU and DEPEN~\cite{dong2009integrating,dong2010global},
correlation fusion (PochampallyCorr)~\cite{pochampally2014fusing}, a mean-field
Ising aggregator over the signed agreement graph (IsingMF), and ClusterVote
(agents grouped by agreement $\ge\theta$, each group casting one vote decided by its
internal majority); and the concurrent LLM aggregators CARE-SVD~\cite{zhao2026care}
and ISP / OW-I~\cite{ai2025beyondmajority}. For LTM, our single-truth implementation
sets each agent's false-positive rate to $(1-\text{sensitivity})/(|V_q|-1)$, which
makes its update identical to one-coin Dawid--Skene; its predictions coincide with
Dawid--Skene on every dataset, so we report one row. CARE is run from the authors'
released code: each (question, option) pair becomes a binary item scored $1$ if the
agent chose it, and the option with the highest CARE score is predicted. Its
structure-learning step needs an invertible judge-correlation matrix; our pools
contain agents with identical answers, which makes the matrix singular, and the
published solver then returns no solution on most splits ($341$ of $520$ solves on
MMLU, all on the other datasets). We therefore report CARE with a small ridge
($10^{-3}I$) added to the correlation matrix, and give the unmodified results in
Appendix~\ref{app:grids}. ISP and OW-I have no released code; we implement them from
the paper's equations, estimating the second-order statistics from answer
co-occurrence. Two single-agent references bound the value of labels: the best agent
selected on the validation half (ties broken at random) and, in
Sec.~\ref{sec:res-labels}, the best agent selected on $k$ labelled questions.
\end{revblock}

\paragraph{Protocol and test of record}
\begin{revblock}
Each dataset is split $40$ times at random into a validation half and a test half.
Methods with a hyperparameter select it on the validation half (grids in
Appendix~\ref{app:grids}) and are scored on the test half; label-free methods are
scored on the test half directly. We report the mean test accuracy over the splits.
\textbf{One test is used for every comparison}: the Nadeau--Bengio corrected
resampled-$t$ over the splits, which accounts for the overlap between splits, with
Holm correction across all comparisons in a table. A method marked ``val.-tuned''
uses the labels of the validation half to pick its hyperparameter and is therefore
not label-free; every BP-TD variant marked ``no labels'' fixes $\theta$ in advance
($0.5$) or sets it with the label-free Otsu rule (Sec.~\ref{sec:res-why}), or does not
use $\theta$ (family). BP-TD is label-blind: its predictions are identical when the
gold labels are shuffled or erased.
\end{revblock}

\paragraph{Reproducibility and data availability}
We release the BP-TD implementation, all baselines, the harvested agent answers
with base-model provenance, the agent prompts and persona texts, and the split
seeds on GitHub (\url{https://github.com/LaureBerti/BPTD}) and as an archived artifact on Zenodo
(DOI:~\href{https://doi.org/10.5281/zenodo.21294747}{10.5281/zenodo.21294747}, all versions;
the version used in this paper is v2.0.0,
\href{https://doi.org/10.5281/zenodo.23916537}{10.5281/zenodo.23916537}),
released under the MIT license. Every number above is recomputable on CPU from the
bundled cached answers, with no model re-querying; we redistribute only the derived
agent answers from the public MMLU, TruthfulQA, SciQ and ARC benchmarks, not the source corpora.

\section{Results}\label{sec:results}
\begin{revblock}
We first compare all methods on the four benchmarks (Sec.~\ref{sec:res-main}), then
take BP-TD apart to see where its gain comes from (Sec.~\ref{sec:res-why}), vary the
agent pool (Sec.~\ref{sec:pools}), and finally look at reliability recovery, scale,
labels, and the regime question (Secs.~\ref{sec:res-rel}--\ref{sec:analysis}).
\end{revblock}

\begin{table*}[t]
\centering
\caption{\rev{Accuracy on the test halves, mean over $40$ splits. $^{+}$/$^{-}$:
significantly above/below majority vote (Nadeau--Bengio corrected resampled-$t$,
Holm-corrected within the dataset). [$n$]: mean number of iterations of loopy
BP-TD; rows marked one-pass use one iteration. CARE, ISP and OW-I were run on the same
splits in a separate pass (Holm family of $7$). Bold: best aggregator per column (the best-agent row uses labels and is excluded).}}
\label{tab:main}
\begin{tabular}{lcccc}
\toprule
Method & MMLU ($160$) & SciQ ($160$) & TruthfulQA ($100$) & ARC ($160$) \\
\midrule
Majority vote (MV) & 0.624 & 0.969 & 0.650 & 0.847 \\
\midrule
BP-TD cooccur, $\theta$ val.-tuned & \textbf{0.751}$^{+}$ {\scriptsize[12]} & 0.969 {\scriptsize[92]} & 0.629 {\scriptsize[14]} & 0.842 {\scriptsize[18]} \\
\quad one-pass = weighted vote & 0.715$^{+}$ & 0.969 & 0.622 & 0.843 \\
BP-TD cooccur, $\theta{=}0.5$ (no labels) & 0.716$^{+}$ & 0.969 & 0.615 & 0.832 \\
BP-TD cooccur, Otsu $\theta$ (no labels) & 0.713$^{+}$ & 0.961 & 0.617 & 0.843 \\
BP-TD family (no labels) & 0.716$^{+}$ {\scriptsize[11]} & \textbf{0.976} {\scriptsize[9]} & 0.586 {\scriptsize[11]} & \textbf{0.879} {\scriptsize[11]} \\
\quad one-pass = weighted vote & 0.717$^{+}$ & \textbf{0.976} & 0.599 & 0.877 \\
BP-TD none ($w_a{\equiv}1$) & 0.513$^{-}$ {\scriptsize[18]} & 0.969 {\scriptsize[7]} & 0.566 {\scriptsize[14]} & 0.849 {\scriptsize[21]} \\
\midrule
ClusterVote, $\theta$ val.-tuned & 0.698 & 0.969 & 0.674 & 0.839 \\
ClusterVote, $\theta{=}0.5$ & 0.672 & 0.969 & 0.653 & 0.826 \\
PochampallyCorr & 0.743 & 0.962 & 0.645 & 0.870 \\
ACCU & 0.530 & 0.969 & 0.562 & 0.850 \\
DEPEN & 0.570 & 0.969 & 0.556 & 0.866 \\
IsingMF (surrogate) & 0.620 & 0.969 & 0.593 & 0.851 \\
\midrule
CARE-SVD (ridge, $\gamma$ val.-tuned) & 0.580 & 0.959 & \textbf{0.683} & 0.859 \\
ISP & 0.639 & 0.969 & 0.526$^{-}$ & 0.853 \\
OW-I & 0.587 & 0.969 & 0.553$^{-}$ & 0.852 \\
\midrule
Dawid--Skene ($=$ LTM here) & 0.500$^{-}$ & 0.969 & 0.551$^{-}$ & 0.840 \\
CRH & 0.558 & 0.969 & 0.570 & 0.849 \\
CATD & 0.493 & 0.965 & 0.555 & 0.844 \\
TruthFinder & 0.525 & 0.969 & 0.554 & 0.858 \\
Investment & 0.506$^{-}$ & 0.969 & 0.554$^{-}$ & 0.817 \\
Sums & 0.619 & 0.969 & 0.590 & 0.848 \\
SSTF-IV & 0.624 & 0.969 & 0.650 & 0.847 \\
SSTF-VI & 0.504$^{-}$ & 0.969 & 0.555 & 0.839 \\
WeightedMV & 0.599 & 0.969 & 0.580 & 0.851 \\
\midrule
Best agent, val.-selected (uses labels) & 0.823$^{+}$ & 0.976 & 0.801$^{+}$ & 0.905 \\
\bottomrule
\end{tabular}
\end{table*}

\begin{revblock}
\subsection{Main comparison}\label{sec:res-main}
Table~\ref{tab:main} gives every method on every benchmark. \textbf{MMLU} is the only
benchmark on which any aggregator significantly beats majority vote, and only
decorrelated BP-TD does so: every BP-TD variant with decorrelation reaches
$0.713$--$0.751$ against $0.624$ for majority vote, including the three variants that
use no labels ($0.713$--$0.716$). The classical estimators and the concurrent
aggregators do not beat majority vote; Dawid--Skene, Investment and SSTF-VI are
significantly below it. Correlation fusion (PochampallyCorr, $0.743$) is numerically
close to BP-TD with $\theta$ tuned ($0.751$); the difference is not significant
($p{=}1.0$ after Holm), and correlation fusion's split-to-split spread is twice
BP-TD's (SD $0.085$ vs.\ $0.046$). We therefore do not claim that BP-TD beats
correlation fusion; we claim that it beats majority vote and the classical
estimators, and that it does so without labels.

On \textbf{SciQ} every method is at or near $0.97$: almost all agents are right, so
there is nothing to correct. On \textbf{TruthfulQA} no aggregator beats majority vote,
decorrelated BP-TD is below it (not significantly), and Dawid--Skene, Investment, ISP
and OW-I are significantly below it. On \textbf{ARC}, BP-TD family is the best label-free
aggregator ($0.879$ vs.\ $0.847$), but the gain is not significant. On all four
benchmarks the best agent selected with the validation labels is the most accurate
method ($0.823$, $0.976$, $0.801$, $0.905$).

\subsection{Where the gain comes from}\label{sec:res-why}
\textbf{Weights, not iteration.} One pass of BP-TD is exactly a $w_a$-weighted vote
(Sec.~\ref{sec:method}), and on every benchmark the one-pass and loopy cooccur rows at
$\theta{=}0.5$ differ by at most $0.007$ ($p>0.75$); for family weights the largest
difference is $0.013$ (TruthfulQA). Iteration adds accuracy only with
$\theta$ tuned on MMLU ($0.751$ vs.\ $0.715$, not significant). The reason is visible
in the weights: at the value usually selected ($\theta{=}0.7$) the strongest agent,
\texttt{qwen2.5:14b}, leaves its group and keeps a full vote, and iterating then raises
the reliability of the strong \texttt{qwen2.5} agents. Without decorrelation,
iteration hurts: BP-TD none falls from $0.624$ (one pass, identical to majority vote)
to $0.513$ after $18$ iterations, because the reliability update rewards the large,
weak \texttt{llama3.2} bloc. The signed reliability smoothing changes full-set MMLU
accuracy by less than $0.013$ in either direction.

\textbf{Soft weights versus hard groups.} At $\theta{=}0.5$ the co-occurrence groups
on MMLU are exactly the base-model families, recovered from the answers alone. With
these identical groups, the soft weighted vote beats ClusterVote, which reduces each
group to one vote decided by its internal majority: $0.716$ vs.\ $0.672$
($+0.043$, $p{=}0.017$). The soft weights keep a split inside a group as fractional
votes and break ties between groups; hard grouping throws both away. On the other
benchmarks the difference is not significant.

\textbf{The threshold $\theta$.} Fig.~\ref{fig:revision}a shows full-set accuracy as a
function of $\theta$. On MMLU the gain is flat at $+0.10$ for $\theta\in[0.4,0.6]$,
peaks at $+0.16$ for $\theta{=}0.7$, and turns negative at $\theta{=}0.9$, where only
near-identical agents are grouped and the weak bloc again outvotes the rest. The label-free Otsu rule, which
splits the pairwise agreement values into two classes, chooses $\theta{=}0.61$ on MMLU
and lands on the plateau. On TruthfulQA most thresholds are below majority vote, and
on ARC and SciQ the curve is flat. A fixed $\theta{=}0.5$ or the Otsu rule loses about
$0.035$ on MMLU relative to tuning $\theta$ on $80$ labels, but tuning on very few
labels is worse than either (Sec.~\ref{sec:res-labels}).

\begin{figure*}[t]
  \centering
  \includegraphics[width=\textwidth]{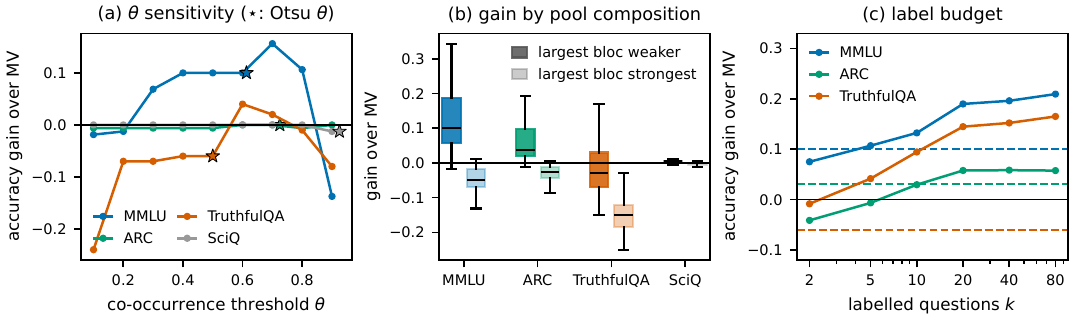}
  \caption{\rev{(a) Gain of loopy BP-TD cooccur over majority vote as a function of
    $\theta$ (full sets); stars mark the label-free Otsu $\theta$. (b) Gain of BP-TD
    family over majority vote across sub-pools. Dark: random sub-pools whose largest
    base-model group is not the most accurate group; light: sub-pools whose largest
    group is the most accurate (random and constructed). (c) Gain of the best agent
    selected on $k$ labelled questions (solid) against label-free BP-TD family (dashed,
    $k{=}0$); mean over $40$ draws.}}
  \label{fig:revision}
\end{figure*}

\subsection{When decorrelation helps: the agent pool}\label{sec:pools}
Our pool has a large, weak \texttt{llama3.2} group and a smaller, strong
\texttt{qwen2.5} group (Table~\ref{tab:agents}), which is the case that
decorrelation favours. To see how much the result depends on this, we evaluated the
label-free methods on $713$ sub-pools per benchmark: $378$ random sub-pools of $5$,
$7$, $9$ and $11$ agents, and $335$ sub-pools built so that the four \texttt{qwen2.5}
agents form the largest group (all four plus up to three \texttt{llama3.2} agents and
any subset of the other three). For each sub-pool we record whether its largest group
is also its most accurate one; this uses the labels only to describe the pool.

Fig.~\ref{fig:revision}b gives the answer. When the largest group is weaker than the
rest, BP-TD family beats majority vote on MMLU in $98\%$ of random sub-pools (median
$+0.10$) and on ARC in $87\%$ (median $+0.04$). When the largest group is the most
accurate, it loses on MMLU in $84\%$ of random and $87\%$ of constructed sub-pools
(median $-0.03$ and $-0.05$) and on ARC in $77\%$ and $94\%$. On TruthfulQA it loses in
most sub-pools of either kind, and on SciQ the gain stays near zero. On MMLU, BP-TD cooccur
($\theta{=}0.5$ or Otsu) gives the same picture; on ARC it gains little in either kind
of sub-pool. The gain is therefore not a
property of the method alone; it is a property of the method and the pool.

A controlled version gives the same picture. We added $k$ noisy copies of the
strongest agent (\texttt{qwen2.5:14b}, $10\%$ of answers replaced by random wrong
options) to the full pool. On MMLU, majority vote rises from $0.612$ to $0.808$ at
$k{=}10$, because the copies are mostly right; BP-TD family stays at $0.71$--$0.72$,
because it discounts the copies as one group; undiscounted BP-TD none reaches $0.825$.
TruthfulQA and ARC behave alike. Decorrelation is the right correction against a
weak dominant bloc and the wrong one against a strong dominant bloc.

\subsection{Reliability recovery}\label{sec:res-rel}
BP-TD estimates a reliability for each agent, and on MMLU these estimates rank agents
by their true accuracy. On the full $160$ questions the Spearman correlation is
$\rho{=}{+}0.94$ (bootstrap over agents, $95\%$ CI $[0.75,1.0]$;
Fig.~\ref{fig:reliability-scatter}), whereas Dawid--Skene, CRH and agreement frequency
give $-0.62$, $-0.62$ and $-0.58$: they rank the weak \texttt{llama3.2} bloc first. On
the $40$ test halves of $80$ questions the correlation is $0.86\pm0.09$ for BP-TD
cooccur ($\theta{=}0.5$; minimum $0.63$) and $-0.65\pm0.05$ for Dawid--Skene. On ARC,
BP-TD family also recovers the ranking ($0.91\pm0.05$). On SciQ and TruthfulQA the
estimates vary as much between splits as their mean, so no ranking can be recovered.
On the MMLU harvests of $5{,}000$ and $10{,}000$ questions, where the half-precision
agents are closer to one another in accuracy, the correlation is lower but still
positive ($0.75$ and $0.68$), and every baseline remains negative. Reliability
estimates are what iteration adds to the weighted vote, and they are the output that
matters when the goal is to choose or audit agents rather than to answer questions.

\begin{figure}[t]
  \centering
  \includegraphics[width=0.86\linewidth]{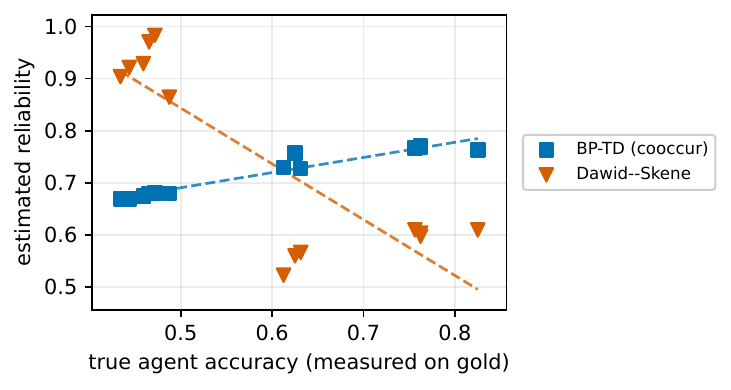}
  \caption{Estimated vs.\ true per-agent reliability on MMLU. BP-TD-cooccur
    recovers the ranking (Spearman $\rho{=}{+}0.94$); Dawid--Skene \emph{inverts}
    it ($\rho{=}{-}0.62$). Dashed lines are least-squares fits.}
  \label{fig:reliability-scatter}
\end{figure}

\subsection{Scale}\label{sec:res-scale}
Table~\ref{tab:scale} repeats the comparison on all $5{,}000$ and $10{,}000$ questions
of the larger MMLU harvests ($10$ splits; each test half has $2{,}500$ or $5{,}000$
questions). The gain over majority vote is smaller than on the $160$-question sample
but significant: $+0.074$ with $\theta$ tuned, and $+0.032$--$0.036$ for BP-TD family
with no labels. BP-TD none stays below majority vote, and the best agent selected on
the validation half is again the most accurate method. Runtime grows close to linearly
with the number of answers: one pass takes $16$, $433$ and $940$\,ms on $160$,
$5{,}000$ and $10{,}000$ questions, and loopy BP-TD needs about $11$ iterations at
scale. On synthetic data up to $n\,m{=}160{,}000$ the fitted runtime exponent is
$0.89$ ($95\%$ CI $[0.79,1.04]$; Fig.~\ref{fig:scalability}), and with $n$ fixed the
cost has the predicted quadratic term in the number of agents.

\begin{table}[t]
\centering
\caption{\rev{MMLU at scale (vLLM, half precision): accuracy on the test halves, mean
over $10$ splits, gain over majority vote in brackets. Marks as in
Table~\ref{tab:main}.}}
\label{tab:scale}
\resizebox{\columnwidth}{!}{%
\begin{tabular}{lcc}
\toprule
Method & $5{,}000$ questions & $10{,}000$ questions \\
\midrule
Majority vote & 0.681 & 0.674 \\
BP-TD cooccur, $\theta$ val.-tuned & 0.755$^{+}$ (+0.074) & 0.748$^{+}$ (+0.073) \\
\quad one-pass = weighted vote & 0.755$^{+}$ (+0.073) & 0.744$^{+}$ (+0.069) \\
BP-TD cooccur, $\theta{=}0.5$ & 0.710 (+0.029) & 0.702$^{+}$ (+0.028) \\
BP-TD cooccur, Otsu $\theta$ & 0.737 (+0.056) & 0.715 (+0.040) \\
BP-TD family & 0.718$^{+}$ (+0.036) & 0.707$^{+}$ (+0.032) \\
BP-TD none & 0.637$^{-}$ (-0.044) & 0.633$^{-}$ (-0.041) \\
ClusterVote, $\theta$ val.-tuned & 0.740$^{+}$ (+0.059) & 0.729$^{+}$ (+0.055) \\
Best agent, val.-selected & 0.790$^{+}$ (+0.109) & 0.787$^{+}$ (+0.113) \\
\bottomrule
\end{tabular}}
\end{table}

\begin{figure}[t]
  \centering
  \includegraphics[width=\linewidth]{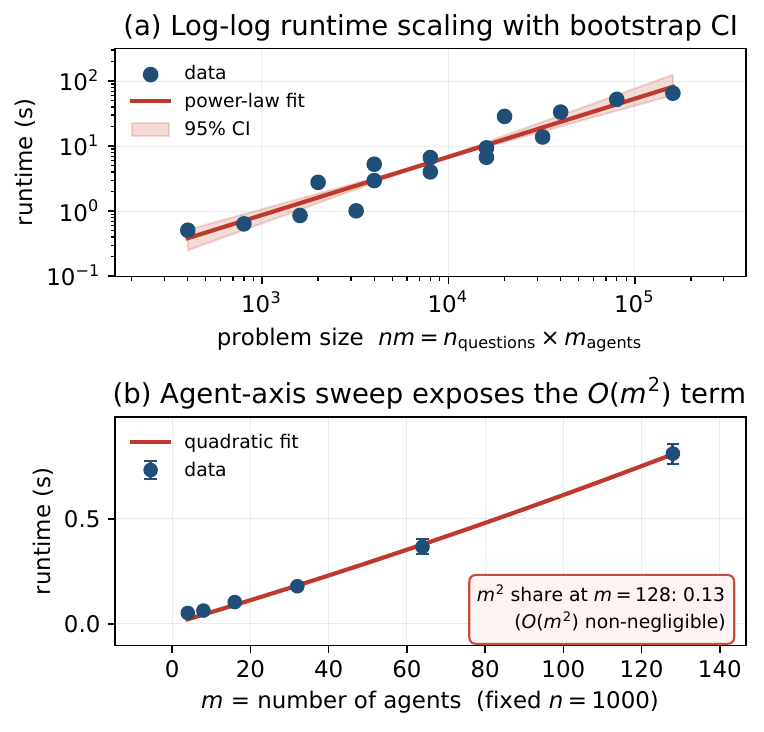}
  \caption{Scalability (synthetic). \textbf{(a)} log--log runtime vs.\ $n\cdot m$;
    \textbf{(b)} agent-axis sweep ($n{=}1000$) exposing the $O(m^2)$ term.}
  \label{fig:scalability}
\end{figure}

\subsection{Labels}\label{sec:res-labels}
\textbf{Label budget.} Fig.~\ref{fig:revision}c compares label-free BP-TD with the
best agent selected on $k$ labelled questions ($40$ random draws; ties broken at
random). The best agent overtakes BP-TD family at $k{=}5$ on MMLU and $k{=}20$ on ARC;
on TruthfulQA it overtakes majority vote at $k{=}5$; on SciQ all differences stay
below $0.015$. Using the same few labels to tune $\theta$ is a poor use of
them: on MMLU, tuning on $2$, $5$ or $10$ labels gives $0.600$, $0.637$ and $0.669$,
below the $0.713$ of the label-free Otsu rule; tuning beats the label-free choice only
from about $40$ labels.

\textbf{Seeding.} Clamping $1$--$20\%$ of the questions to their gold answers changes
the accuracy on the remaining questions by at most $0.001$ on MMLU and SciQ, $0.01$ on ARC and
$0.02$ on TruthfulQA ($40$ draws; SD up to $0.03$). With $13$ agents each seed adds
one observation per agent to the reliability counts, which is too little to move them.

\textbf{Do the classical estimators work at all?} On synthetic data with $13$
independent agents of accuracy $0.30$--$0.92$, Dawid--Skene (and LTM) beat majority
vote ($0.992$ vs.\ $0.978$ on $160$ questions, $0.992$ vs.\ $0.973$ on $1{,}000$; $20$
seeds), as does BP-TD none. Their collapse on MMLU is therefore caused by the data:
Dawid--Skene's answers agree on $98.8\%$ of questions with the plurality of the six
\texttt{llama3.2} agents, which is right only $47\%$ of the time. On the same
independent agents, BP-TD cooccur loses a little ($0.983$), the price of discounting
agents that agree only because they are right.

\end{revblock}

\begin{figure}[t]
  \centering
  \includegraphics[width=0.86\linewidth]{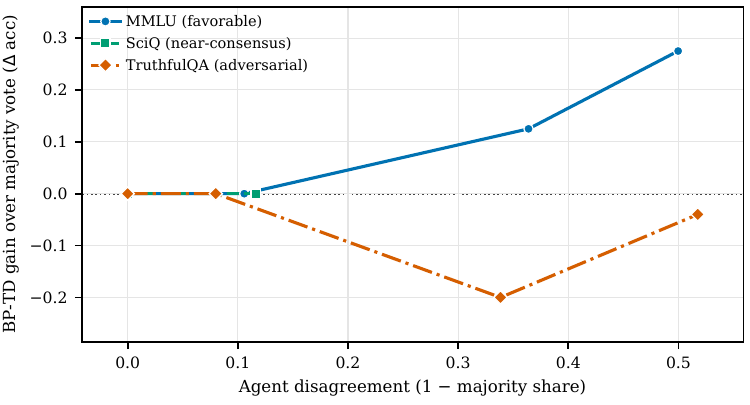}
  \caption{BP-TD's accuracy gain over majority vote vs.\ per-question agent
    disagreement ($1-$majority share), stratified into quartiles \emph{within}
    each dataset (single held-out split). The gain rises with disagreement on MMLU,
    is non-positive throughout on TruthfulQA, and is flat on SciQ.}
  \label{fig:gain-disagreement}
\end{figure}

\begin{figure*}[t]
  \centering
  \includegraphics[width=0.92\textwidth]{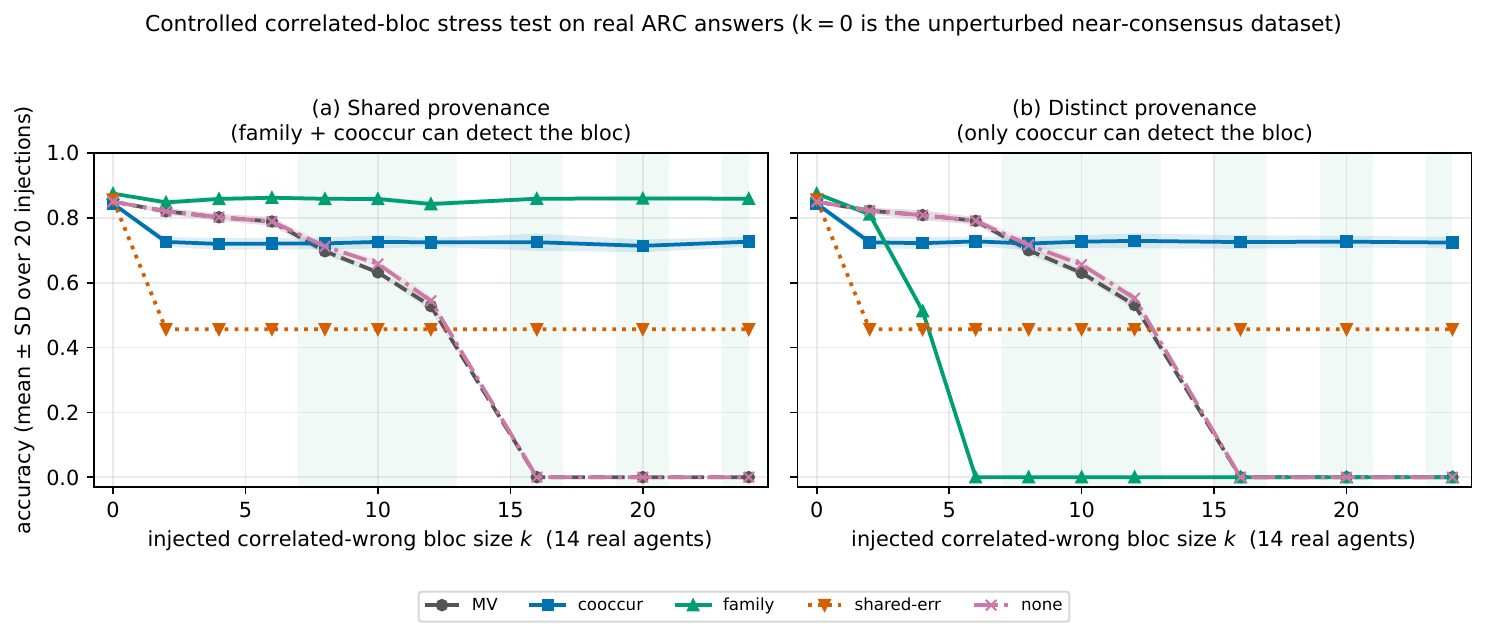}
  \caption{Controlled correlated-bloc stress test on the real ARC answers:
    we append $k$ synthetic agents sharing one wrong answer per question and sweep
    $k$ at fixed hyperparameters, under (a)~shared and (b)~distinct bloc provenance
    ($\uparrow$ better). Curves: majority vote (MV) and four BP-TD variants (cooccur,
    family, shared-err, none). Mean\,$\pm$\,SD over $20$ injections.}
  \label{fig:bloc-stress}
\end{figure*}

\subsection{When does agent truth discovery work?}\label{sec:analysis}
\begin{revblock}
The sub-pool study gives the practical rule: decorrelation helps when the agents that
agree most are not the agents that are most often right. Two further analyses refine it.

\textbf{Disagreement.} Within MMLU, the gain over majority vote grows with
per-question disagreement ($+0.00$, $+0.00$, $+0.125$, $+0.275$ across quartiles on one
held-out split; Fig.~\ref{fig:gain-disagreement}); within TruthfulQA it is never
positive; within SciQ it is flat. MMLU and TruthfulQA have nearly the same mean signed
agreement ($0.33$ and $0.35$), so how much agents agree does not decide the outcome;
whether their shared answer points toward or away from the truth does.

\textbf{A wrong bloc.} On the ARC answers we appended $k$ synthetic agents that all
give the same wrong answer (Fig.~\ref{fig:bloc-stress}). Majority vote and BP-TD none
collapse once the bloc is the majority. BP-TD family is unaffected when the bloc shares
a base model ($\approx$$0.86$ at every $k$); when the bloc hides its provenance,
BP-TD cooccur still recovers the truth ($\approx$$0.72$). The truth-aware
\texttt{shared\_error} weights fail throughout ($\approx$$0.46$), because the bloc
defines the current truth estimate and so appears to share no errors. This is the
mirror image of the strong-bloc experiment of Sec.~\ref{sec:pools}: the same discount
that rescues the answer from a wrong bloc throws away a right one.

\textbf{No label-free diagnostic.} A practitioner needs to know which case applies
without labels. We tested two candidate statistics, chosen before seeing the results,
on the $2{,}852$ sub-pools: the agreement between the largest group's plurality answer
and the plurality answer of the other agents, and the fraction of questions on which
BP-TD changes majority vote's answer. Neither separates sub-pools where BP-TD helps
from those where it hurts: the AUC is $0.53$--$0.67$ per benchmark for the first and
$0.48$--$0.85$ for the second, and $0.53$ and $0.51$ when the benchmarks are pooled.
Telling the two cases apart remains the main open problem; where it cannot be done,
spending a few labels on choosing an agent is the safer option.
\end{revblock}

\section{Discussion and Limitations}\label{sec:disc}
\begin{revblock}
\textbf{What the gain means in practice.} Consider a pipeline that answers $10{,}000$
MMLU-style questions with these $13$ agents and no labels. BP-TD family answers about
$320$ more questions correctly than majority vote ($0.707$ vs.\ $0.674$) and reports
which agents are reliable; its one-pass form ($0.702$) runs in under a second on CPU. If the pool's
largest group is its strongest, the same choice costs about $300$--$500$ answers per
$10{,}000$ (median $-0.03$ to $-0.05$ on MMLU sub-pools). If $20$ questions can be
labelled, choosing the best agent is better still: on the $160$-question MMLU sample
it reaches $0.805$, against $0.711$ for BP-TD family. BP-TD is therefore a tool for the label-free case, for pools
in which redundant variants of a weak model would otherwise outvote stronger,
independent models, and for auditing agent reliability.

\textbf{Limitations.} (i)~\emph{Benchmarks and scale.} The regime results rest on four
benchmarks, of which only MMLU was harvested at scale; SciQ and TruthfulQA were not.
All agents are local open models of up to $14$B parameters; larger API-served models
were not tested and may be less redundant. (ii)~\emph{Pool composition.} The headline
gain depends on a pool whose largest group is weak (Sec.~\ref{sec:pools}). (iii)~\emph{No
diagnostic.} We found no label-free way to tell whether decorrelation will help.
(iv)~\emph{Dependence model.} The weights are a heuristic tempering of the likelihood,
not a generative model of dependence, and the truth-aware \texttt{shared\_error}
weights fail when the dependent bloc is the majority. (v)~\emph{Contamination.} The base
models have likely seen these benchmarks. Every aggregator receives the same answers,
so contamination does not favour one aggregator on a given answer matrix, but it can
change the answer matrix itself: if models of one family memorized the same items,
their errors are more shared, which is the variable our regime analysis depends on.
The TruthfulQA result in particular may partly reflect contamination rather than the
benchmark's adversarial design alone.
\end{revblock}

\section{Conclusion}\label{sec:conclusion}
\begin{revblock}
Truth discovery among LLM agents fails in its classical form because agents built on
one base model agree for reasons unrelated to the truth. Discounting such agents,
as BP-TD does with provenance or co-occurrence weights inside a Dawid--Skene model,
recovers a significant gain over majority vote on MMLU without labels, at $160$ and
at $10{,}000$ questions, and recovers the ranking of agents by accuracy where
classical estimators invert it. Its first iteration is a weighted vote; iteration
contributes the reliability estimates. Across thousands of sub-pools, the gain appears
when the largest correlated group is weaker than the other agents and reverses when
it is the strongest, and we found no label-free statistic that predicts which case
holds. With twenty labelled questions, choosing the best single agent is better than
every aggregator we tested. Two directions follow: a label-free test for the
direction of shared errors, and a rule for switching from aggregation to agent
selection as labels arrive.
\end{revblock}

\section*{Acknowledgment}
\rev{The author used DeepL as a writing-assistance tool for translating and
enhancing the text and for \LaTeX{} proofreading, and Claude for automating the experiments.
All scientific contributions---experimental design, algorithm development, formalization, data
collection and analysis, result interpretation, and the claims made in the paper---are entirely the
author's own work. The AI assistant did not generate any experimental results, derive any
mathematical content, or contribute to the conceptual framing of the research.}

\appendices
\section{Hyperparameter grids and CARE details}\label{app:grids}
\begin{revblock}
Tuned methods select their hyperparameter on the validation half by accuracy and
freeze it before scoring the test half. Grids: BP-TD cooccur
$\theta\in\{0.2,\dots,0.8\}$ (step $0.1$); ClusterVote $\theta\in\{0.3,\dots,0.7\}$;
TruthFinder $\gamma\in\{0.1,0.3,0.5,1.0\}$; CARE-SVD $\gamma\in\{0.1,0.2,0.25,0.5,
0.75,1,2,3,5,7,10\}$ (the authors' grid). The thresholds most often selected for
BP-TD cooccur are $0.7$ on MMLU, $0.2$ on SciQ, $0.5$ on TruthfulQA and $0.8$ on ARC.
Label-free BP-TD uses $\theta{=}0.5$ or the Otsu threshold of the off-diagonal signed
agreements (mean $0.56$ over MMLU test halves). Tuning the iteration count of
Dawid--Skene and LTM, CATD's $\alpha$ and TruthFinder's $\gamma$ on the validation half
does not lift them above majority vote on MMLU (Dawid--Skene $0.501$, CATD $0.493$,
LTM $0.500$, TruthFinder $0.525$).

\textbf{CARE without the ridge.} With the published code unchanged, the structure
solve fails on $341$ of $520$ solves on MMLU and on all $520$ on SciQ, TruthfulQA and
ARC, and CARE then returns majority vote. Its accuracy is $0.598$ ($\gamma{=}1$) and
$0.607$ ($\gamma$ tuned) on MMLU, and equal to majority vote on the other three
benchmarks. On synthetic independent agents the ridge version is slightly below
majority vote ($0.959$ vs.\ $0.978$), which suggests that the one-indicator-per-option
encoding we use for multiple choice is not where CARE is strongest.
\end{revblock}

\section{BoolQ pilot}\label{app:boolq}
\begin{revblock}
Before the main harvest we ran the same pool on $128$ BoolQ questions. BoolQ asks a
yes/no question about a passage, but only the two RAG agents received a passage; the
other $11$ agents answered from memory. The pilot therefore does not test closed-book
aggregation, and we do not use it in the regime analysis. Under the protocol of
Sec.~\ref{sec:setup} ($40$ splits), no aggregator differs significantly from majority
vote ($0.682$): ClusterVote $0.722$, BP-TD cooccur $0.706$, BP-TD family $0.681$,
Dawid--Skene $0.654$, PochampallyCorr $0.640$. The best agent selected on the
validation half reaches $0.826$, significantly above majority vote.
\end{revblock}

\bibliographystyle{IEEEtran}

\end{document}